\documentclass{article}

\usepackage[final]{neurips_2021}
\makeatletter
\renewcommand{\@noticestring}{NeurIPS 2021 Workshop on Bayesian Deep Learning}
\makeatother

\usepackage[utf8]{inputenc} 
\usepackage[T1]{fontenc}    
\usepackage{hyperref}       
\usepackage{url}            
\usepackage{booktabs}       
\usepackage{amsfonts}       
\usepackage{nicefrac}       
\usepackage{microtype}      
\usepackage{xcolor}         
\usepackage[pdftex]{graphicx}  
\usepackage[english]{babel}
\usepackage{subcaption}

\usepackage{amsmath}

\newcommand{\argmin}{\mathop{\mathrm{argmin}}}

\newcommand{\argmax}{\mathop{\mathrm{argmax}}}

\title{Generation of data on discontinuous manifolds via continuous stochastic non-invertible networks }

\author{Mariia Drozdova$^{1,2}$\thanks{M. Drozdova and S. Voloshynovskiy are corresponding authors.}, Vitaliy Kinakh$^1$, Guillaume Quétant$^{1,2}$, \\ \textbf{Tobias Golling$^2$ \& Slava Voloshynovskiy$^1$} \\
  $^1$Department of Computer Science\\
  $^2$Department of Particle Physics\\
  University of Geneva\\
  Switzerland\\
  \texttt{\{mariia.drozdova,svolos\}@unige.ch} \\
}

\begin{document}

\maketitle

\begin{abstract}
The generation of discontinuous distributions is a difficult task for most known frameworks such as generative autoencoders and generative adversarial networks. Generative non-invertible models are unable to accurately generate such distributions, require long training and often are subject to mode collapse. Variational autoencoders (VAEs), which are based on the idea of keeping the latent space to be Gaussian for the sake of a simple sampling, allow an accurate reconstruction, while they experience significant limitations at generation task. In this work, instead of trying to keep the latent space to be Gaussian, we use a pre-trained contrastive encoder to obtain a clustered latent space. Then, for each cluster, representing a unimodal submanifold, we train a dedicated low complexity network to generate this submanifold from the Gaussian distribution. The proposed framework is based on the information-theoretic formulation of mutual information maximization between the input data and latent space representation. We derive a link between the cost functions and the information-theoretic formulation. We apply our approach to synthetic 2D distributions to demonstrate both reconstruction and generation of discontinuous  distributions using continuous stochastic networks.

\end{abstract}

\section {Introduction}
\label{introduction}




The generation of data with discontinuous distributions with non-invertible networks represents a great interest for many problems in high energy physics, astrophysics and chemistry all dealing with high dimensional data. The previous attempts to develop generative models for discontinuous distributions show limited performance of GANs \cite{dumoulin2017adversarially}\cite{goodfellow2014generative}\cite{karras2019stylebased} and VAE models \cite{kingma2014autoencoding}. Flow models \cite{dinh2017density} can handle this problem to some extent but they face the complexity issues when the dimensionality of data increases. Hybrid models such as SurVAE \cite{nielsen2020survae} try to solve this problem by a combination of non-invertible and invertible networks based on Flows.

In this paper, we present a new information-theoretic stochastic contrastive generative adversarial network SC-GAN. The SC-GAN is a hybrid system that is based on a deterministic encoder producing an interpretable latent space and a stochastic decoder representing a generator. The generator architecture is a set of fully connected layers implemented based on a stochastic EigenGAN network \cite{He2021} conditioned on a set of random noise vectors at each layer. The model is trained both in the {\it{reconstruction}} mode (with fixed noise vectors) and in the {\it{generative}} mode.
The contrastive encoder is trained independently of the generator. The latent space of the encoder is then clustered using K-Means and approximated by many low-complexity mapping networks which try to shape Gaussians into corresponding cluster distributions. Finally, the decoder is trained jointly for reconstruction and generation using the likelihood with the corresponding discriminators.

We provide an information-theoretical interpretation of the proposed model in Section \ref{formulation}. In Section \ref{experiments}, we perform the analysis of both auto-encoding mode and generation of toy 2D datasets: Eight Gaussians, Checkerboard, Two spirals, Abs, Sinewaved cube and Four circles. 

\section {Information-theoretic formulation}
\label{formulation}

The proposed framework is schematically shown in Figure 1 and consists of three stages of training. The reconstruction information terms in stages 2 and 3 are variational lower bounds on mutual information under the corresponding encoder-induced joint distributions.

\subsection{The training of the encoder (stage 1)}

The encoder training is motivated by maximizing the mutual information between the data ${\bf X}$ and its latent space representation ${\bf E}$: 
\begin{equation}
\label{MI_SIMCLR}
\hat{{\phi}_{\varepsilon}} = \argmax_{{\phi}_{\varepsilon}} I_{{\phi}_{\varepsilon}}({\bf X} ;{\bf E}), 
\end{equation}
where $I_{{\phi}_{\varepsilon}}({\bf X} ;{\bf E}) =  \mathbb{E}_{p_{\phi_\varepsilon}(\mathbf{x},\varepsilon)}\left[   \log \frac{q_{ {\phi}_{\varepsilon}}({ \varepsilon} | {\bf x} )}{q_{ {\phi}_{\varepsilon}}({ \varepsilon})}   \right]$. Here, $p_{\mathrm{data}}(\mathbf{x})$ denotes the data distribution, $q_{\phi_\varepsilon}(\varepsilon|\mathbf{x})$ the conditional distribution induced by the encoder, and $\phi_\varepsilon$ the encoder parameters. The corresponding joint and marginal distributions are $p_{\phi_\varepsilon}(\mathbf{x},\varepsilon)=p_{\mathrm{data}}(\mathbf{x})q_{\phi_\varepsilon}(\varepsilon|\mathbf{x})$ and $q_{\phi_\varepsilon}(\varepsilon)=\int p_{\mathrm{data}}(\mathbf{x})q_{\phi_\varepsilon}(\varepsilon|\mathbf{x})\,d\mathbf{x}$, respectively.

\begin{figure}[ht] 
  \begin{subfigure}[b]{0.5\linewidth}
    \centering
    \includegraphics[width=1.0\linewidth]{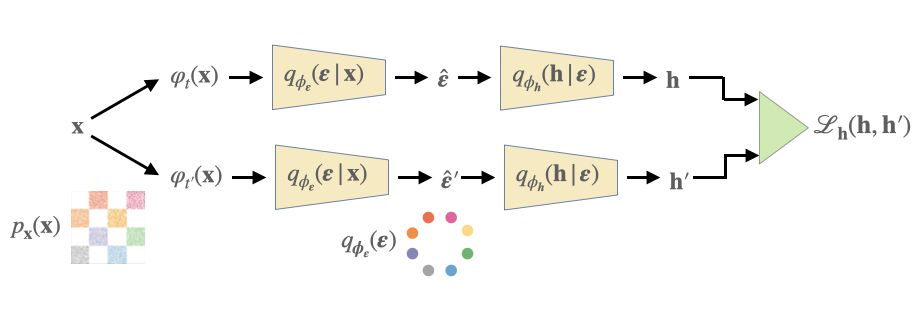} 
    \caption{Stage 1} 
    \label{fig7:a} 
    \vspace{4ex}
  \end{subfigure}
  \begin{subfigure}[b]{0.5\linewidth}
    \centering
    \includegraphics[width=1.0\linewidth]{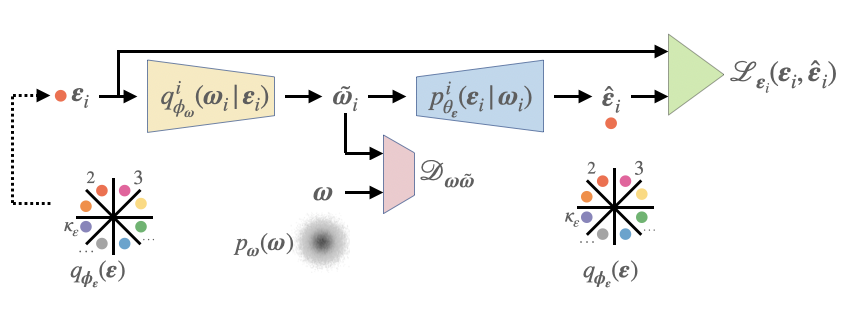} 
    \caption{Stage 2} 
    \label{fig7:b} 
    \vspace{4ex}
  \end{subfigure} 
  \begin{subfigure}[b]{0.5\linewidth}
    \centering
    \includegraphics[width=1.0\linewidth]{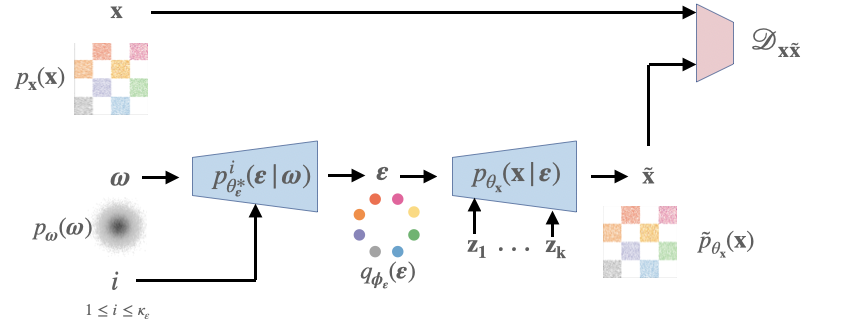} 
    \caption{Stage 3 : generation step} 
    \label{fig7:c} 
  \end{subfigure}
  \begin{subfigure}[b]{0.5\linewidth}
    \centering
    \includegraphics[width=1.0\linewidth]{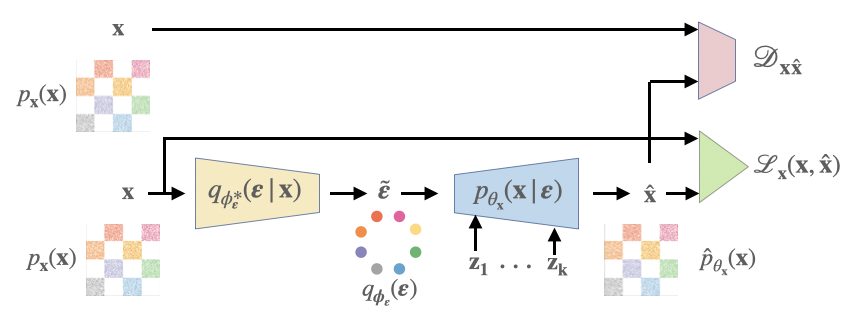} 
    \caption{Stage 3 : auto-encoding step} 
    \label{fig7:d} 
  \end{subfigure} 
  \caption{The proposed framework: Stage 1 - training of the encoder, Stage 2- training of the mapping network, Stage 3 - training of the decoder for the simultaneous reconstruction and generation.}
  \label{fig:proposed scheme} 
\end{figure}

 The encoder is trained independently from the decoder using contrastive losses (Figure \ref{fig7:a}). The maximization can be considered in the scope of the InfoNCE framework \cite{oord2019representation} and technically implemented using for example SimCLR contrastive learning \cite{chen2020simple}. For our toy datasets we choose simple augmentations based on the addition of small noise to the input data.

%
%
%
%
%
%

\begin{figure}
  \begin{subfigure}{0.15\textwidth}
  \centering
    \includegraphics[width=\linewidth]{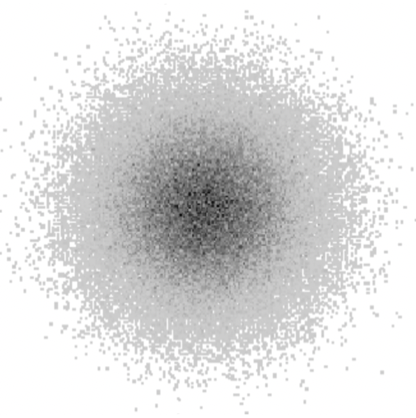}
    \caption{a single Gaussian pdf} \label{fig:mapping network:1a}
  \end{subfigure}%
  \hspace*{\fill}   
  \begin{subfigure}{0.15\textwidth}
  \centering
    \includegraphics[width=\linewidth]{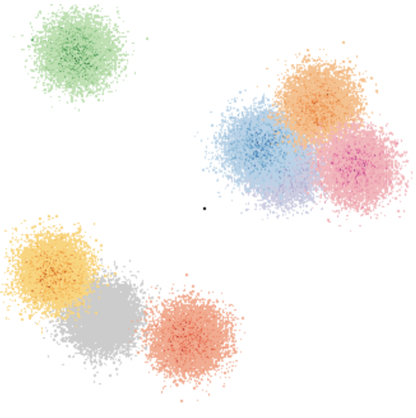}
    \caption{cluster centers with the Gaussian noise} \label{fig:mapping network:1c}
  \end{subfigure}
  \hspace*{\fill}
   \begin{subfigure}{0.15\textwidth}
   \centering
    \includegraphics[width=\linewidth]{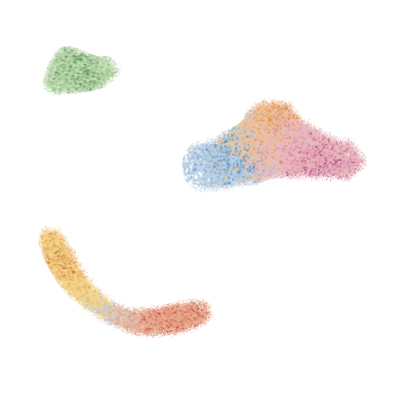}
    \caption{mapping network} \label{fig:mapping network:1d}
  \end{subfigure}%
  \hspace*{\fill}   
  \begin{subfigure}{0.15\textwidth}
  \centering
    \includegraphics[width=\linewidth]{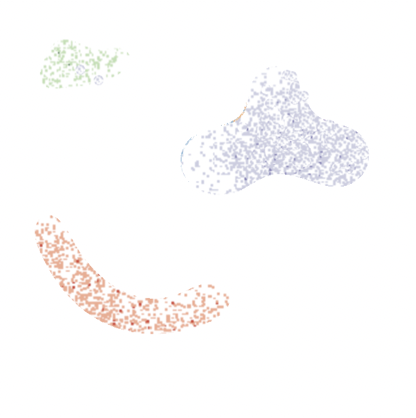}
    \caption{Subdataset partition} \label{fig:mapping network:1e}
  \end{subfigure}%
  \hspace*{\fill}   
  \begin{subfigure}{0.15\textwidth}
  \centering
    \includegraphics[width=\linewidth]{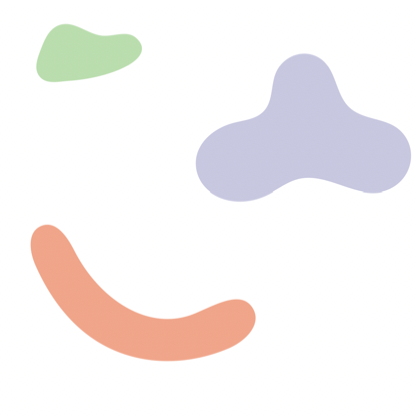}
    \caption{real data - continuous manifold} \label{fig:mapping network:1f}
  \end{subfigure}

\caption{All explored ways to model the latent space. Last columns is the ground truth data.} \label{fig:mapping network}
\end{figure}

\begin{figure}
\centering
    \includegraphics[width=0.8\linewidth]{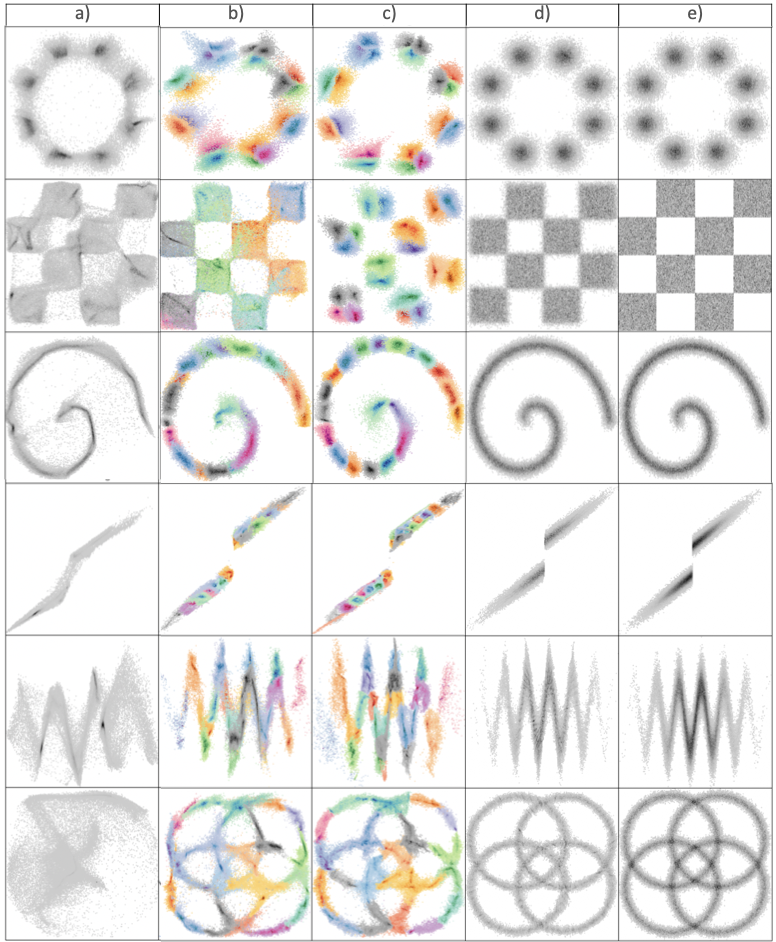}
    \caption{The generation results from different latent space models : a) from a single Gaussian pdf, b) from cluster centers with the Gaussian noise, c) from the mapping network, d) from real data. e) is a target distribution.}
\label{fig:generation results}
\end{figure}

\subsection{ The training of the latent space mapping networks (stage 2)}

The mapping networks aim at generating the complex latent space $ \mathbf \varepsilon$ from a simple distribution $p_{ \omega} ( \omega)$ (Fig. \ref{fig7:b}). 
The latent space $ \varepsilon$ has a discontinuous clustered nature following the distribution $q_{\phi_{\varepsilon}} ( \mathbf\varepsilon)$. 
A simple continuous MLP mapper cannot generate complex $q_{ \phi_{ \varepsilon}} (\mathbf\varepsilon)$ from a unimodal $p_{\omega} (\omega)$. 
For this reason, we consider splitting $q_{ \phi_{ \varepsilon}} (\mathbf\varepsilon)$ on a set of unimodal sub-distributions in such a way that a simple MLP model can be used to generate each unimodal sub-distribution. 
We use a simple K-means \cite{1056489} clustering to produce ${K}_{\mathbf \varepsilon}$ distinct subsets. 
Then for all ${ \mathbf \varepsilon} \in \mathcal{C}_{i}$, $i = 1, ..., K_{\mathbf \varepsilon}$, we trained adversarial auto-encoders (AAE) \cite{makhzani2016adversarial} with the latent space $p_{\omega} (\omega)$ following the Gaussian distribution. 
Alternatively, one can use Flows to construct a mapper from 
$p_{\omega} (\omega)$ 
to 
$q_{\phi_{\varepsilon}} (\mathbf\varepsilon)$. 
Each AAE is defined by a pair of the encoder $q^{i}_{\phi_{ \omega}} (\omega|\mathbf\varepsilon)$ and the decoder $p^{i}_{\theta_{\varepsilon}} (\mathbf\varepsilon| \mathbf\omega)$ for each cluster ${i} \in {1, ..., K_{\mathbf\varepsilon}}$. 
The training of the AAE is motivated by the optimization problem:
\begin{equation}
\label{AAE_train_loss}
\max_{\phi_\omega^i,\theta_\varepsilon^i}I^{i}_{{\phi}_{ \omega}}({\bf{E}} ; {\bf{W}}) + \lambda_{\varepsilon}I^{i}_{{  \phi}_{ \omega},{ \theta}_{\mathbf \varepsilon}}({{\bf{W}} ; {\bf {E}}}), \end{equation}
where 
\begin{equation*}\begin{split}  I^{i}_{{  \phi}_{  \omega}}({\bf E} ; {\bf W}) & = \mathbb{E}_{q^{i}_{{  \phi_{  \mathbf \varepsilon}} (  \mathbf\varepsilon)}}\left[ \mathbb{E}_{q^{i}_{{  \phi_{  \omega}} (  \omega |   \mathbf\varepsilon)}}\left[\log\frac{q^{i}_{  \phi_{  \omega}} (  \omega |   \mathbf\varepsilon)}{q^{i}_{  \phi_{  \omega}} (  \omega)} \times \frac{p_{  \omega} (  \omega)}{p_{  \omega} (  \omega)}\right]\right] \\
& =
\mathbb{E}_{q^{i}_{{  \phi_{  \mathbf \varepsilon}} (  \mathbf\varepsilon)}}\left[\mathbb{D}_{\mathrm{KL}}(q^{i}_{  \phi_{  \omega}} (  \omega | {\bf {E}} =
\mathbf\varepsilon) || p_{  \omega} (  \omega))\right] - \mathbb{D}_{\mathrm{KL}}(q^{i}_{  \phi_{  \omega}} (  \omega ) || p_{  \omega} (  \omega)),\end{split}\end{equation*}
and 
\begin{equation}  I^{i}_{{  \phi}_{  \omega},{  \theta}_{{\mathbf \varepsilon}}}({{\bf W} ; {\bf E}}) = \mathbb{E}_{q^{i}_{{  \phi_{  \mathbf \varepsilon}} (  \mathbf\varepsilon)}}\left[ \mathbb{E}_{q^{i}_{{  \phi_{  \omega}} (  \omega |   \mathbf\varepsilon)}}\left[\log\frac{p^{i}_{\theta_{  \varepsilon}} (  \mathbf\varepsilon|  \mathbf\omega)}{q^{i}_{\phi_{  \varepsilon}} (  \mathbf\varepsilon)}\right]\right] = H_{\phi^{i}_{  \mathbf \varepsilon}}({\bf {E}}) - H_{\phi^{i}_{  \mathbf \omega}, \theta_{\mathbf \varepsilon}}({\bf {E}} | {\bf {W}}),\end{equation}
where 
$  H_{\phi^{i}_{  \mathbf \varepsilon}}({\bf {E}}) = - \mathbb{E}_{q^{i}_{{  \phi_{  \mathbf \varepsilon}} (  \mathbf\varepsilon)}}\left[\log{q^{i}_{\phi_{  \varepsilon}} (  \mathbf\varepsilon)}\right]$, 
$  H_{\phi^{i}_{  \mathbf \omega}, \theta_{\mathbf \varepsilon}}({\bf {E}} | {\bf {W}}) = - \mathbb{E}_{q^{i}_{{  \phi_{  \mathbf \varepsilon}} (  \mathbf\varepsilon)}}\left[ \mathbb{E}_{q^{i}_{{  \phi_{  \omega}} (  \omega |   \mathbf\varepsilon)}}\left[\log{p^{i}_{\theta_{  \varepsilon}} (  \mathbf\varepsilon|  \mathbf\omega)}\right]\right]$ 
and 
$\lambda_{\varepsilon}$ 
is a constant for the importance of each term. 
As $\mathbb{E}_{q^{i}_{{  \phi_{  \mathbf \varepsilon}} (  \mathbf\varepsilon)}}\left[\mathbb{D}_{\mathrm{KL}}(q^{i}_{  \phi_{  \omega}} (  \omega | {\bf {E}}=  \mathbf\varepsilon) || p_{  \omega} (  \omega))\right] \geq 0$ and $H_{\phi^{i}_{  \mathbf \varepsilon}}({\bf {E}})$ does not depend on the parameters of the networks, maximizing a lower
bound on the objective in Eq.~\eqref{AAE_train_loss}
is equivalent to minimizing:
$(\hat{  \phi}^{i}_{  \omega},\hat{  \theta}^{i}_{  \mathbf \varepsilon} ) = \argmin_{ {  \phi}^{i}_{  \omega},{  \theta}^{i}_{  \mathbf \varepsilon}} \mathbb{D}_{\mathrm{KL}}(q^{i}_{  \phi_{  \omega}} (  \omega ) || p_{  \omega} (  \omega)) + \lambda_{\varepsilon}H_{\phi^{i}_{  \mathbf \omega}, \theta_{\mathbf \varepsilon}}({\bf {E}} | {\bf {W}}),$
which has the structure of an AAE objective \cite{makhzani2016adversarial}.
  
\subsection{ The training of the decoder (stage 3)}

The decoder training for both \textit{reconstruction} and \textit{generation} modes is motivated by the following information-theoretic optimization problem: 
\begin{equation}
\label{decoder_loss}
\max_{\theta_{\mathrm x}} I_{{  \phi}^{*}_{  \mathbf\varepsilon}, {  \theta}_{\mathrm x}}({\bf E} ; {\bf X}) + \lambda_{\mathrm x}I_{{  \theta}^{*}_{  \mathbf\varepsilon}, {  \theta}_{\mathrm x}}({\bf E} ; {\bf X}), 
\end{equation}
where the reconstruction mode corresponds to the term:
\begin{equation*}\begin{split}
I_{{  \phi}^{*}_{  \mathbf\varepsilon}, {  \theta}_{\mathrm x}}({\bf E} ; {\bf X}) & = \mathbb{E}_{p_{\bf x}({\bf x})}\left[\mathbb{E}_{q_{\phi^{*}_{  \mathbf \varepsilon}}(  \mathbf \varepsilon|  {\bf x} )} \left[ \log\frac{p_{\theta_{\bf x}}({\bf x} |   \mathbf \varepsilon)}{p_{{\bf x}}({\bf x})}\times\frac{\hat{p}_{\theta_{\bf x}}({\bf x})}{\hat{p}_{\theta_{\bf x}}({\bf x})} \right] \right] \\ & = - H_{\phi^{*}_{  \mathbf \varepsilon}, \theta_{\mathrm x}}({\bf {X}} | {\bf {E}}) - \mathbb{D}_{\mathrm{KL}}(p_{{\bf {x}}}({\bf {x}} ) || \hat{p}_{\theta_{\bf x}}({\bf x} )) + H(p_{{\bf {x}}}({\bf {x}} ) ; \hat{p}_{\theta_{\bf x}}({\bf x} ))\end{split}\end{equation*} 
with 
$H_{\phi^{*}_{\mathbf\varepsilon},\theta_{\mathrm x}}
({\bf X}|{\bf E})
=
-\mathbb{E}_{p_{\bf x}({\bf x})}
\left[
\mathbb{E}_{q_{\phi^{*}_{\mathbf\varepsilon}}
(\mathbf\varepsilon|{\bf x})}
\left[
\log p_{\theta_{\bf x}}({\bf x}|\mathbf\varepsilon)
\right]
\right]$, $\mathbb{D}_{\mathrm{KL}}(p_{{\bf {x}}}({\bf {x}} ) || \hat{p}_{\theta_{\bf x}}({\bf x} )) =  \mathbb{E}_{p_{\bf x}({\bf x})}\left[\log\frac{p_{\bf x}({\bf x})}{\hat{p}_{\theta_{\bf x}}({\bf x})}\right]$ and $H(p_{{\bf {x}}}({\bf {x}} ) ; \hat{p}_{\theta_{\bf x}}({\bf x} )) = - \mathbb{E}_{p_{\bf x}({\bf x})}\left[\log{\hat{p}_{\theta_{\bf x}}({\bf x})}\right]$ and the generation mode corresponds to the term:
\begin{equation*}
\begin{split}
I_{\theta^*_{\mathbf\varepsilon},\theta_{\mathrm x}}
({\bf E};{\bf X})
&=
\mathbb{E}_{p_\omega(\omega)}
\left[
\mathbb{E}_{p_{\theta^*_{\mathbf\varepsilon}}
(\mathbf\varepsilon|\omega)}
\left[
\mathbb{E}_{p_{\theta_{\bf x}}
({\bf x}|\mathbf\varepsilon)}
\left[
\log
\frac{p_{\theta_{\bf x}}({\bf x}|\mathbf\varepsilon)}
{\tilde p_{\theta_{\bf x}}({\bf x})}
\right]
\right]
\right]
\\
&=
\mathbb{E}_{p_{\theta^*_{\mathbf\varepsilon}}
(\mathbf\varepsilon)}
\left[
\mathbb{D}_{\mathrm{KL}}
\left(
p_{\theta_{\bf x}}({\bf x}|{\bf E}=\mathbf\varepsilon)
\|
\tilde p_{\theta_{\bf x}}({\bf x})
\right)
\right].
\end{split}
\end{equation*}  with
$  p_{  \theta^{*}_{  \mathbf \varepsilon}}(  \mathbf \varepsilon) = \mathbb{E}_{p_{   \omega}({  \omega})}\left[p_{\theta^{*}_{  \mathbf \varepsilon}}(  \mathbf \varepsilon|    \omega )\right]$ and $\mathbb{D}_{\mathrm{KL}}(p_{{\bf {x}}}({\bf {x}} ) || \tilde{p}_{\theta_{\bf x}}({\bf x} )) = \mathbb{E}_{p_{\bf x}({\bf x})}\left[\log\frac{p_{\bf x}({\bf x})}{\tilde{p}_{\theta_{\bf x}}({\bf x})}\right]$ and $\lambda_{\mathrm x}$ is a constant controlling the trade-off between the two terms, $\hat{p}_{{\bf {x}}}$ and $\tilde{p}_{{\bf {x}}}$ denote the distributions of reconstruction and generated data, respectively. 
To additionally enforce matching of the reconstructed and generated data distributions with the real data distribution, we adopt the following surrogate training objective:
\begin{equation*}
\label{decoder_loss_reduced}
\hat{  \theta}_{\mathrm x} = \argmin_{{  \theta}_{\mathrm x}} H_{\phi^{*}_{  \mathbf \varepsilon}, \theta_{\mathrm x}}({\bf {X}} | {\bf {E}}) + \mathbb{D}_{\mathrm{KL}}(p_{{\bf {x}}}({\bf {x}} ) || \hat{p}_{\theta_{\bf x}}({\bf x} )) + \lambda_{\mathrm x}\mathbb{D}_{\mathrm{KL}}(p_{  {{\bf x}}} ({\bf x}) || \tilde{p}_{\theta_{\bf x}}({\bf x})).
\end{equation*}

\section{Experiments and Conclusions}
\label{experiments}

\begin{table}[ht]
\begin{center}
	\begin{tabular}{c || c | c | c | c }
		\hline
		
		    & single Gaussian pdf & noisy cluster centers & mapping network & real data\\ 
		\hline\hline
		Eight Gaussians      & 2.339  & 0.053  & 0.065 & 0.016  \\
		\hline
		Checkerboard    & 0.337  & 0.034  & 0.015 & 0.009    \\
		\hline
		Two Spirals     & 1.771  & 0.058  & 0.062 & 0.011   \\
		\hline
		Abs    & 0.133  & 0.021  & 0.029 & 0.019 \\
		\hline
		Sinewaved cube & 0.062  & 0.023  & 0.024 & 0.021  \\
		\hline
		Four circles  & 0.031  & 0.034  & 0.044 & 0.035
\end{tabular}
\caption{The reconstruction results for the different latent space models: latent vector sampled from a single Gaussian pdf, from cluster centers with Gaussian noise, from the mapping network output and from random training data latents.}
\label{Tab:Tcr}
\end{center}
\end{table}

We perform generation experiments on 2D datasets using different ways for latent space modeling (the Fig. \ref{fig:mapping network}). In the first setting $  \mathbf \varepsilon$ is sampled from the Gaussian probability density function. Then we cluster the $  \mathbf \varepsilon$-space using K-Means. In the second setting we place the Gaussian in the cluster centers and use this as input. Finally, we train an individual network for each cluster to shape the Gaussian closer to the real shape of the cluster (stage 2 of the training). For generation we can also use $  \mathbf \varepsilon$ from the subset used to train the encoder. This case is an extreme case when the number of clusters is equal to the size of the dataset.  We show the results of the generation in Fig. \ref{fig:generation results}.

Covering the latent space with the clusters which are approximated by simple fully connected layers leads to state-of-the-art results (the fourth column in Fig. \ref{fig:generation results}) for the generative models which are not based on INNs. 

We show the Mean Square Error in Table \ref{Tab:Tcr} to demonstrate reconstruction error. We fix the noise vectors and we take the latent vector directly from the output of the encoder from Stage 1. The training stage of generation is different: we take the latent vector from a single Gaussian pdf, from cluster centers with the Gaussian noise, from the mapping network and from real data. We notice that despite poor generation, network can still perform good results in reconstruction with simplest latent space modeling (one Gaussian) which is a sign of overfitting. It is also interesting to note that the better generation is, the worth are the results of reconstruction and vice versa. Modeling the latent space as discontinuous allow us to marry the mode of reconstruction and generation. In the limit case with number of clusters equal to the number of points in the train set we get the best results.

\begin{ack}
This research was partially funded by the SNF Sinergia project (CRSII5-193716): Robust Deep Density Models for High-Energy Particle Physics and Solar Flare Analysis (RODEM). The authors are thankful to Johnny Raine and Sebastian Pina-Otey for their feedback on the paper and discussion. 
\end{ack}

{
\small

\bibliographystyle{unsrt}
\bibliography{egbib}
}

\appendix



\end{document}